\documentclass{article}

\usepackage{times}
\usepackage{geometry}
\usepackage{hyperref}
\usepackage{url}
\usepackage{graphicx}
\usepackage{multirow}
\usepackage{booktabs}       
\usepackage{amsfonts}       
\usepackage{amsmath}
\usepackage{amsthm}
\usepackage{amssymb}
\usepackage{pifont}
\usepackage{natbib}
\title{ProbaNet: Proposal-balanced Network for Object Detection}
\date{}

\author{Jing Wu, Xiang Zhang\thanks{Corresponding author}, Mingyi Zhou, Ce Zhu\\
	\\
	\textit{University of Electronic Science and Technology of China}\\
	{\tt\small uestchero@std.uestc.edu.cn}
}

\begin{document}
	
	\maketitle
	
	\begin{abstract}
		Candidate object proposals generated by object detectors based on convolutional neural network (CNN) encounter easy-hard samples imbalance problem, which can affect overall performance. In this study, we propose a Proposal-balanced Network (ProbaNet) for alleviating the imbalance problem. Firstly, ProbaNet increases the probability of choosing hard samples for training by discarding easy samples through threshold truncation. Secondly, ProbaNet emphasizes foreground proposals by increasing their weights. To evaluate the effectiveness of ProbaNet, we train models based on different benchmarks. Mean Average Precision (mAP) of the model using ProbaNet achieves 1.2$\%$ higher than the baseline on PASCAL VOC 2007. Furthermore, it is compatible with existing two-stage detectors and offers a very small amount of additional computational cost.
	\end{abstract}
	
	\section{Introduction}
	\label{sec:intro}
\begin{figure}
		\begin{center}
			\includegraphics[width=0.75\textwidth]{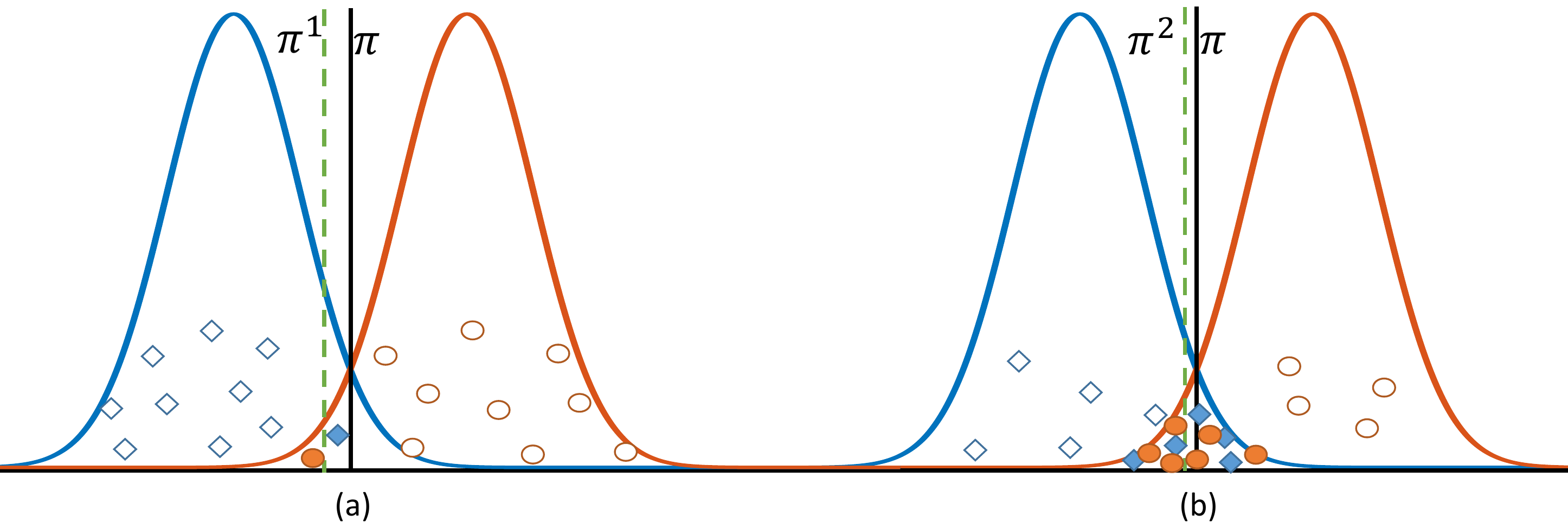}
		\end{center}
		\caption{Illustration of easy-hard imbalance problem after the random sampling strategy in detectors. It is assumed that the positive (blue) and negative (orange) samples satisfy the Gaussian distribution. The hollow circles, solid circles, hollow diamonds and solid diamonds denote the easy positive samples, hard positive samples, easy negative samples and hard negative samples, respectively. $\pi$ is the real decision surface, $\pi ^ 1$ and $\pi ^ 2$ are the learned decision surface.}
		\label{fig:imb}
	\end{figure}

	Current mainstream object detectors are two-stage detectors represented by the R-CNN series (\emph{e.g}, R-CNN~\cite{Girshick2014RichFH}, Fast R-CNN~\cite{girshick2015fast}, Faster R-CNN~\cite{Ren2015FasterRT} and R-FCN~\cite{pang2019libra}) and one-stage detectors such as RetinaNet~\cite{lin2017focal}, YOLO~\cite{Redmon2016YouOL,Redmon2017YOLO9000BF} and SSD~\cite{Liu2016SSDSS,fu2017dssd}. A two-stage detector applies the convolutional network to classify the candidate object proposals generated in the first stage, while a one-stage detector classifies a dense set of object locations directly. However, whether it is a two-stage detector or a one-stage detector, they are facing easy-hard samples imbalance problem. The number of easy negatives far exceeds the number of hard negatives. This within-class imbalance problem has an adverse effect on detection performance. As shown in Figure \ref{fig:imb}, hard samples (overlapping areas of two categories) have more contributions to build the decision surface in training. For a network with insufficient hard samples to train (Figure \ref{fig:imb} (a)), if the number of easy samples reduces, the probability of hard samples to be selected will be raised under the sampling strategy. The learned decision surface will be more close to the real decision surface (Figure \ref{fig:imb} (b)).
	
	To some extent, the foreground-background class imbalance problem is addressed in the R-CNN series detectors as the majority background proposals are filtered out through sampling heuristics. 
	In two-stage detectors, region proposals which are more likely to be background samples generated in proposal stage are rapidly decreased. The second classification stage uses the sampling strategy to fix the ratio (1:3) of foreground-background candidate proposals in a mini-batch for training, which aims to guarantee a manageable balance between foreground and background samples. However, the random sampling in two-stage detectors cannot solve easy-hard samples imbalance problem, because the distribution of samples in IoU (Intersection over Union) is uneven~\cite{pang2019libra}. On the contrary, a one-stage detector has to handle a dense set of candidate object locations, even applying sampling strategy to it, it is still dominated by plenty of easy background samples during the training procedure. Commonly used and effective methods to balance the easy-hard samples are OHEM (online hard sample mining~\cite{Shi2015EarlyBD,li2017s}), Focal loss~\cite{lin2017focal} and so forth~\cite{Wang2017AFastRCNNHP}. They either create hard samples or perform hard samples mining in the training stage.
	
	In this paper, we propose a Proposal-balanced Network (ProbaNet) collaborated with RPN (Region Proposal Network) to address the easy-hard samples imbalance problem in current two-stage object detectors. ProbaNet assigns lower weights to majority background, which include many easy samples, and higher weights to object areas, which include many meaningful proposals and hard samples. Then we utilize a threshold to filter out easy samples, thus the distribution of training data will be more balanced than its original form. Another advantage of ProbaNet is that it tends to assign higher focus to candidate proposals that are more meaningful (\emph{e.g}, the proposals containing objects), thus more meaningful proposals will be selected in the test stage. In addition, we apply a statistical constraint in ProbaNet which guarantees a larger difference among proposals, thus enabling ProbaNet to distinguish the hard samples. ProbaNet has good compatibility and is easy to be embedded in other deep network models. Experimental results confirm that the proposed ProbaNet alleviates the easy-hard samples imbalance problem in object detectors and improves the object detection strength while offering very small extra computational cost.
	
	\section{Related work}
	\label{sec:work}
    \textbf{Object detectors:}
	Deep learning based detectors dominate modern object detection. R-CNN~\cite{Girshick2014RichFH} generates candidate object locations that contain all objects theoretically in the first stage, uses CNN to extract features for every region proposal and classifies them as foreground or background classes through SVM (Support Vector Machine) algorithm in the second stage. Fast R-CNN~\cite{girshick2015fast} achieves near real-time rates using the convolutional network for extracting features and classifying, regardless of the time spent on region proposals. Faster R-CNN~\cite{Ren2015FasterRT} using RPN to integrate the region proposal process and the second stage into a single and unified network. The other mainstream modern object detectors are one-stage detectors (\emph{e.g}, SSD~\cite{Liu2016SSDSS,fu2017dssd} and YOLO~\cite{Redmon2016YouOL,Redmon2017YOLO9000BF}). These detectors have been adjusted for speed, but their accuracy usually lags behind the two-stage approach. 
	Multiple improvements in terms of speed and accuracy have been proposed for object detectors or backbones. For sample, for object location, G-CNN~\cite{Najibi2016GCNNAI} removes the object proposal phase in the CNN-based object detection framework and models the detection problem as an iterative regression problem which further improves accuracy and speed. Instead of correcting the object positions by regression in Faster R-CNN, Grid R-CNN~\cite{Lu2018GridR} drops the fully connected layers in the original model, instead, it uses a fully convolutional network (FCN~\cite{long2015fully}) to get more precise object positions.
	For the capability of detection, Wang \emph{et al}~\cite{Wang2017AFastRCNNHP} trained a model with samples having occlusions and deformations generated by the adversarial network to make detectors invariant to occlusions and deformations, while Li \emph{et al}~\cite{Li2017PerceptualGA} improved small object detection through a generative adversarial network.
	The popular method attention mechanisms~\cite{Chorowski2015AttentionBasedMF} have utility across many tasks including object detections, such as~\cite{Ren2015FasterRT,Hu2018SqueezeandExcitationN}. In terms of the base network structure, Hu \emph{et al}~\cite{Hu2018SqueezeandExcitationN} proposed the Squeeze-and-Excitation (SE) block that adaptively recalibrates channel-wise feature responses which brought outstanding improvements for CNNs. Deformable convolution and deformable RoI pooling proposed by Dai \emph{et al}~\cite{dai2017deformable} enhance the geometric transformation modeling capability of CNNs.
	
	\noindent
	\textbf{Class imbalance:}
	Class imbalance problem has been thoroughly studied in classical machine learning and data mining. In the field of object detection, R-CNN-like detectors address foreground-background class imbalance through sampling strategy and a cascade classifier~\cite{Ren2015FasterRT}, but they are still facing with easy-hard samples imbalance problem. Both two-stage detectors and one-stage detectors address classification loss and box regression loss imbalance by applying the certain penalty on the corresponding loss~\cite{girshick2015fast,Ren2015FasterRT,Redmon2016YouOL,Redmon2017YOLO9000BF}. OHEM~\cite{Shi2015EarlyBD,li2017s}, Focal Loss~\cite{lin2017focal} and GHM (Gradient Harmonizing Mechanism)~\cite{li2018gradient} unearth hard samples according to their loss. OHEM is sensitive to noise samples and Focal loss needs extra hyper-parameters for data having a different distribution, while GHM performs worse than Focal loss in small object detection. Wang \emph{et al}~\cite{Wang2017AFastRCNNHP} created hard positive samples through an adversarial network. Libra R-CNN~\cite{pang2019libra} propose three components named as IoU-balanced sampling, balanced feature pyramid and balanced $L1$ loss to solve imbalance problem from sample level, feature level, and objective level, respectively.
	
	\section{ProbaNet: Approach Details}
	\label{sec:ProbaNet}
	
	Two-stage object detectors address the foreground-background class imbalance problem through fixing the ratio of foreground to background samples (1:3). Because the number of easy background proposals far exceeds than the number of hard negatives, random sampling strategy is likely to choose more easy negatives under uneven distribution in IoU (Intersection over Union). These easy negatives have less contribution to the update of the gradient. So far, most scholars have focused on creating or mining hard samples through the loss function and balancing the loss between classification and regression based on cost-sensitive methods~\cite{zhou2006training}. There have been some attempts to assign weights to the datasets themselves. Can we assign weights to different candidate proposals and raise the probability of hard negatives to be selected? In this paper, we attempt to meet this objective by proposing ProbaNet. The procedure of the proposed ProbaNet is discussed in the following subsections.
	
	\begin{figure}
		\begin{center}
			\includegraphics[width=.92\textwidth]{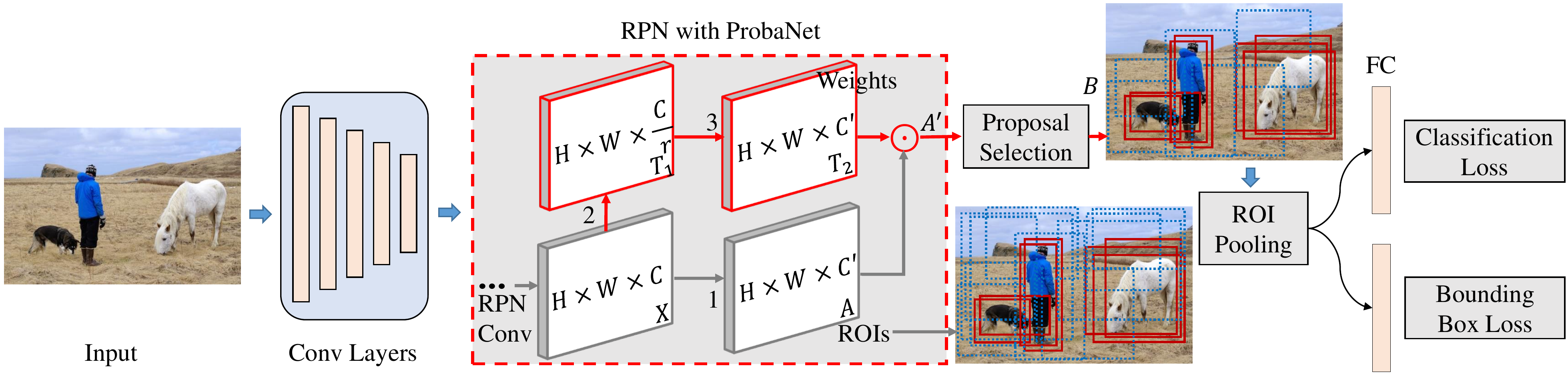}
		\end{center}
		\caption{Faster R-CNN with ProbaNet. It is still a single and unified network. The red dotted frame is the detail of the black solid frame. (Point 1,2 and 3 represent convolutional transformation.)}
		\label{fig:teaser}
	\end{figure}
	
	\subsection{ProbaNet Design}
	
	An overview of Faster R-CNN with ProbaNet framework is shown in Figure \ref{fig:teaser}. We embed ProbaNet in RPN, so it is still a single and unified network. The shaded part is RPN with ProbaNet, the upper and left part (the red arrow flow) in the dotted box represents details of ProbaNet, the rest is just the original components in Faster R-CNN defined in~\cite{Ren2015FasterRT}. As discussed above, in the field of object detection, usually, the proportion of background is larger than objects, even overwhelming foreground. Just looking at the image shown in left-bottom of Figure \ref{fig:teaser}, the black dog, the person with a blue coat and the white horse have less proportion than the grassland background. Taking proposals generated in RPN as a sample, for the training stage, the majority proposals containing no objects (easy samples) are easy to be classified and have limited contribution to the detector in training. For the test stage, the number of proposals containing no objects far exceeds the proposals containing objects which potentially harm the overall performance of the model.
	
	It is a well-known fact that the objective of RPN in Faster R-CNN is to generate proposals. So can we further improve the network capability to reasonably weight different proposals to address the imbalance problem? We propose ProbaNet aimed at suppressing samples having a small contribution for training or deserving high focus for test. The architecture and formulation of ProbaNet are discussed as follows:
	
	As is shown in the dotted box of Figure \ref{fig:teaser}, following the convolutional layers in RPN, we stack two convolutional layers and a sigmoid activation to generate the corresponding weights with the proposals. Suppose that convolutional feature maps generated from the backbone are represented as $\mathbf{X} \in \mathbb{R}^{H \times W \times C}$, where $H \times W$ is the spatial dimension and $C$ is the channel of $\mathbf{X}$ . $\mathbf{A} \in \mathbb{R}^{H\times W \times {C^{'}}}$ represents the candidate proposals generated during RPN and $C^{'}$ is the channel of $\mathbf{A}$. For simplicity, we ignore the ReLU function following convolution layers here. Then the process of ProbaNet can be seen as
	
	\begin{align}
	\label{ProbaNet_function}
	\begin{split}
	\mathbf{A}^{'}=\mathbf{A} \odot \mathbf{T}_2 
	& = F^1_{conv}(\mathbf{X}) \odot \sigma(F^3_{conv}(\mathbf{T}_1)) \\
	& = F^1_{conv}(\mathbf{X}) \odot \sigma(F^3_{conv}(F^2_{conv}(\mathbf{X}))).
	\end{split}
	\end{align}
	
	Here, ${\mathbf{T}_1} \in {\mathbb{R}^{H \times W \times {C / r}}}$ represents temp feature maps during ProbaNet, $r$ is the reduction parameter that needs to be mannually set, and ${\mathbf{T}_2} \in {\mathbb{R}^{H \times W \times C}}^{'}$ stands for the final weights learned for $\mathbf{A}$. ${\mathbf{A}^{'}} \in {\mathbb{R}^{H \times W \times {C^{'}}}}$ represents the final candidate object locations with weights. ${F^k_{conv}}$ ($k=1,2,3$) is taken to be convolutional transformation, $ \sigma $ is Sigmoid function where $ \sigma (x) = {1 / (1 + {e^{ - x}})}$. $\odot$ is Hadamard product. We use $1 \times 1$ sliding window in both two convolution layers because ProbaNet focuses on assignment of appropriate weights to candidate proposals. This enables ProbaNet to establish associations among candidate proposals.
	
	\subsection{Proposal Selection}
	
	\textbf{For training stage:} Due to the effect of minimizing the loss in ProbaNet based RPN, proposals generated by RPN working together with ProbaNet will get more proper focus than those only generated by RPN. However, the easy negatives with slight weights but vast numbers still have higher chances to be selected under the random sampling strategy. They have less contribution to models in the training stage. To choose more valuable samples (hard samples) in the training stage, we further dismiss proposals with weights less than $th$ $(0 \le th < 1)$ . Therefore the final proposals $B$ reserved are
	
	\begin{equation}
	\label{final_proposals}
	\mathbf{B}(i,j) =
	\begin{cases} 
	\mathbf{A}^{'}(i,j),  & \mbox{if } \mathbf{T}_2(i,j) \succ th \\
	0, & \mbox{otherwise.} 
	\end{cases}
	\end{equation}
	
	Thus more easy samples will be screened out in training. In equation \ref{final_proposals}, ${i,j} \in {\mathbb{N}}$ and $i,j \leq H$, $i,j \leq W$. The value of parameter $th$ should be considered according to the distribution of datasets. If the easy-hard samples imbalance is heavy, $th$ should be set small enough because the ProbaNet will give little focus on proposals which are likely to be easy negatives samples. On the contrary, $th$ should be set higher if the class imbalance is light. The rest network components are the same as the original Faster R-CNN.
	
	Furthermore, For better assisting the ProbaNet, we impose a variance constraint on proposals to strengthen the discrimination capacity of ProbaNet before its focus assignment,
	\begin{equation}
	\label{variance}
	V =  \mathrm{E}\left[(\mathbf{X}-\mathrm{E}\left[{\mathbf{X}}\right])^T (\mathbf{X}-\mathrm{E}\left[{\mathbf{X}}\right])\right],
	\end{equation}
	where $\mathrm{E}\left[{\mathbf{X}}\right] $ is the average value of candidate proposals' weights $\mathbf{X}$, which are the input of the ProbaNet. Equation \ref{variance} can enhance the discrimination between weights of foreground and background. In our experiments, we utilize this constrain as loss function as follows,
	\begin{equation}
	\label{variance_loss}
	\mathcal{L}_{ProbaNet} = \beta e^{1/V}, 
	\end{equation}
	Where $\mathcal{L}_{ProbaNet}$ is the loss function of the proposed ProbaNet. The $\beta$ guarantees $\mathcal{L}_{ProbaNet} < \mathcal{L}_{cls}$, where $\mathcal{L}_{cls}$ is the classification loss of RPN and $\beta$ is automatically adjusted in training.
	
	\noindent
	\textbf{For test stage:} The process of ProbaNet in the test is the same with its training process except the $th$ in equation \ref{final_proposals} is set to 0. Our experiments show that the proposed ProbaNet assign higher weights to the proposals containing objects and give them higher probabilities of being selected. 
	
	In general, we propose RPN with ProbaNet to address within-class imbalance problem. The black solid frame in Figure \ref{fig:teaser} shows how ProbaNet works. ProbaNet gives a certain focus on proposals generated by RPN. Through the threshold truncation and the variance constraint, the number of easy negatives will be limited such that to alleviate the adverse effects brought by easy-hard samples imbalance problem in training. In addition, the proposals containing objects with higher weights are more likely to be selected.
	
	\subsection{Analysis of Model Complexity}
	
	\begin{table}
		\begin{center}
			\footnotesize
			\begin{tabular}{|l|cc|cc|}  
				\hline
				&Re-FastR  &Re-FastR+ProbaNet  &Re-R-FCN  &Re-R-FCN+ProbaNet\\
				\hline  \hline
				Params(MB)      &522.91   &522.98   &216.33  &216.40\\
				FLOPs(G)        &151.30   &151.33   &75.75   &75.78\\
				\hline
			\end{tabular}
		\end{center}
		\caption{Compare the re-implementation model with the model with ProbaNet. Re-FastR is short for re-implemented Faster R-CNN. The parameter $r$ is 16 and $th$ is 0.5. Size of the input is $600\times800\times3$.}
		\label{tab:paflop}
	\end{table}
	
	For practical use, object detectors with the proposed ProbaNet should offer a good trade-off between improved performance and increased model complexity. We consider a comparison between the re-implementation model and the ProbaNet based model. Both experiments are conducted on a server with one NVIDIA GTX-1080Ti GPU.
	We calculate the additional parameters introduced by ProbaNet. The additional parameters are generated solely by the two convolutional layers, making up a small portion of the total network capacity. Concretely, the total number of extra parameters is given by
	
	\begin{equation}
	\label{extra_cost}
	C \times (\frac{C}{16}+1) + \frac{C}{16} \times(C^{'}+1),
	\end{equation}
	where $C$ refers to the dimension of the input channels and $C^{'}$ denotes the dimension of the output channels. As shown in Table \ref{tab:paflop}, ProbaNet introduces only extra 0.07MB parameters and 0.03G FLOPs. Faster R-CNN takes 44.9ms per image, compared to 45.2ms for Faster R-CNN with ProbaNet, while the latter get 1.1$\%$ higher mAP than the former. Hence, it is evident that ProbaNet improves the object detector capabilities while offering a very little amount of extra computational cost.
	
	\section{Experiments}
	\label{sec:exp}
	
	In this section, we conduct experiments on the PASCAL VOC 2007 and PASCAL VOC 2012 detection benchmarks to investigate the effectiveness of the proposed ProbaNet based method. We perform the most ablative study on PASCAL VOC 2007 and PASCAL VOC 2012 with Faster R-CNN. In addition, we compare ProbaNet with OHEM, Focal loss and compare R-FCN with R-FCN+ProbaNet. Furthermore, we visualize some weights learned by ProbaNet and show proposals with top5$\%$ weights in the original image. Finally, we validate that ProbaNet indeed alleviates the easy-hard samples imbalance problem.
	
	\subsection{Implementation Details}
	
	We use identical training schemes for the re-implemented models and models with ProbaNet. We normalize the input images by averaging channel subtraction and re-scale the input so that the shorter side has 600 pixels. In RPN, we sample 256 proposals from 2 images within a mini-batch. The whole network is trained with stochastic gradient descent (SGD). All models are trained with an initial learning rate of 0.001, weight decay of 0.005 after every 5 epoch and momentum of 0.9. We apply ProbaNet to the whole image during the test stage.
	
	\subsection{Experimental settings}
	Experiments are performed using VGG-16 for Faster R-CNN and ResNet-101 for R-FCN. The VGG-16 and ResNet-101 models are pre-trained on ImageNet. For the PASCAL VOC datasets, we use the 'trainval' set for training and 'test' set for testing. PASCAL VOC 2007 dataset consists of 5k trainval images composed of 13K objects and 5k test images over 20 object categories, while about 12k trainval images composed of 27K objects and 11k test images over 20 object categories in PASCAL VOC 2012 dataset, which has no public labels and requires the use of the evaluation server. Furthermore, we train models with the 'trainval' set in both VOC datasets. Among these experiments, the only difference between the re-implementation method and our proposed method is that the proposed method uses ProbaNet.
	
	\subsection{Experiments on PASCAL VOC}
	In this subsection, we validate the effectiveness of the proposed ProbaNet based method by taking into account different baseline architectures. In the first experiment, we develop Faster R-CNN and compare its performance with ProbaNet based Faster R-CNN, the results are reported in Table \ref{tab:07test}. In the second experiment, we develop Faster R-CNN using OHEM or Focal loss and compare its performance with ProbaNet based Faster R-CNN, the results are reported in Table \ref{tab:focal}.  Additionally, we also developed another baseline architecture i.e. R-FCN and compared its performance with ProbaNet based R-FCN, the results are shown in Table \ref{tab:07test}. In the third experiment, we evaluate the sensitivity of the hyper-parameter $th$ and $r$ inside the proposed ProbaNet, the results can be seen in Table \ref{tab:th_r}. Details about these experiments are discussed as follows:
	
	\newcommand{\tabincell}[2]{\begin{tabular}{@{}#1@{}}#2\end{tabular}}
	
	\begin{table}
		\begin{center}
			\footnotesize
			\begin{tabular}{|l|c|ccc|cc|}  
				\hline
				\multirow{2}{*}{\tabincell{c}{Test\\ data}}  &\multirow{2}{*}{\tabincell{c}{Training\\ data}}   &\multicolumn{3}{c}{FastR}   &\multicolumn{2}{|c|}{R-FCN}\\
				\cline{3-5} \cline{6-7}  &  &FastR   &Re-FastR  &Re-FastR+ProbaNet  &Re-R-FCN  &Re-R-FCN+ProbaNet\\
				\hline \hline
				\multirow{3}{*}{07}   &07     &69.9  &69.9  &\textbf{71.1}   &74.1   &\textbf{74.9}\\
				&12     &-     &73.2  &\textbf{73.5}   &76.2   &\textbf{76.3}\\
				&07+12  &73.2  &75.9  &\textbf{76.4}   &79.1   &\textbf{79.5}\\
				\hline
				\multirow{3}{*}{12}   &07     &-     &63.5  &\textbf{64.4}   &69.0     &\textbf{69.1}\\
				&12     &67.0  &69.9  &\textbf{70.2}   &73.2     &\textbf{73.6}\\
				&07+12  &70.4     &72.0  &\textbf{72.3}   &75.1     &\textbf{75.5}\\
				\hline
			\end{tabular}
		\end{center}
		\caption{Detection results (mAP($\%$)) on PASCAL VOC test set. The backbone is VGG-16. FastR is short for the official Faster R-CNN, Re-FastR represents for re-implementated Faster R-CNN. Training data: ``07'': VOC 2007 trainval, ``07+12'': union set of VOC 2007 trainval and VOC 2012 trainval. ``12'': VOC 2012 trainval. The data having underline means that it is trained with the union set of VOC 2007 trainval and VOC 2012 trainval.}
		\label{tab:07test}
	\end{table}
	
	\noindent
	\textbf{Experiments with Faster R-CNN and R-FCN.} As is shown in Table \ref{tab:07test}, Faster R-CNN combined with ProbaNet gives $0.3\% \sim 1.2\%$ boost upon the baseline, while R-FCN combined with ProbaNet gives $0.8\%$ improvement on the baseline. 
	
	\begin{table}
		\begin{center}
			\footnotesize
			\begin{tabular}{|l|cccc|}  
				\hline
				Training data   &Re-FastR    &Re-FastR+OHEM   &Re-FastR+Focal loss  &Re-FastR+ProbaNet\\
				\hline  \hline
				07        &69.9   &70.2   &70.4   &\textbf{71.1}   \\
				12        &73.2   &73.0   &73.4   &\textbf{73.5}   \\
				07+12     &75.9   &75.4   &76.2   &\textbf{76.4}   \\
				\hline
			\end{tabular}
		\end{center}
		\caption{Detectors' detection results(mAP($\%$)) on PASCAL VOC 2007 test set when combined with different methods. Re-FastR is short for re-implementated Faster R-CNN. The backbone for Faster R-CNN is VGG-16 and for R-FCN is ResNet-101. Training data: ``07'': VOC 2007 trainval, ``07+12'': union set of VOC 2007 trainval and VOC 2012 trainval. ``12'': VOC 2012 trainval.}
		\label{tab:focal}
	\end{table}
	
	\noindent
	\textbf{Experiments with methods for easy-hard samples imbalance.} Table \ref{tab:focal} shows results comparing ProbaNet with other methods for addressing easy-hard samples imbalance in object detection. For Faster R-CNN trained with VOC 2007 trainval set, using OHEM and Focal loss achieve 70.2$\%$ and 70.4$\%$ mAP, respectively. When combined with ProbaNet, mAP of the model has been improved up to 71.1$\%$. When trained with VOC 2012 and the union set of VOC 2007 and VOC 2012, all these methods have similar boost upon the baseline.
	
	\begin{table}
		\begin{center}
			\footnotesize
			\begin{tabular}{|l|c|ccccc|}  
				\hline
				\multirow{2}{*}{$th=0.5$} &$r$    &16            &8     &4     &2     &1\\
				\cline{2-7}         &mAP($\%$)    &\textbf{71.1} &70.1  &70.6  &70.5  &70.5\\
				\hline \hline
				\multirow{2}{*}{$r=16$}  &$th$    &0.3  &0.4   &0.5            &0.6     &0.7\\
				\cline{2-7}         &mAP($\%$)    &70.5 &70.2  &\textbf{71.1}  &70.6    &70.1\\
				\hline
			\end{tabular}
		\end{center}
		\caption{Detection results on VOC 2007 test set using Faster R-CNN with ProbaNet. Training with VOC 2007 trainval set and the backbone is VGG-16.}
		\label{tab:th_r}
	\end{table}
	
	\noindent
	\textbf{Evaluating the sensitivity of the hyper-parameters.} Table \ref{tab:th_r} shows how the hyper-parameters $th$ and $r$ in ProbaNet effect the models' performance. $r$ is decided to control the ability to represent interdependencies among proposals, $th$ is meant to select which proposals need to be filtered out. To evaluate the sensitivity of this hyper-parameters in ProbaNet, we conduct experiments with ProbaNet based Faster R-CNN using VOC 2007 dataset. We observe that with enough focus, ProbaNet is able to overfit to the interdependencies among proposals. For VOC 2007 dataset, the threshold $th = 0.5$ and $r = 16$ achieves the best performance.
	
	\subsection{Experiments to further analyze effectiveness of ProbaNet}
	
\begin{figure}
		\begin{center}
			\includegraphics[width=0.8\textwidth]{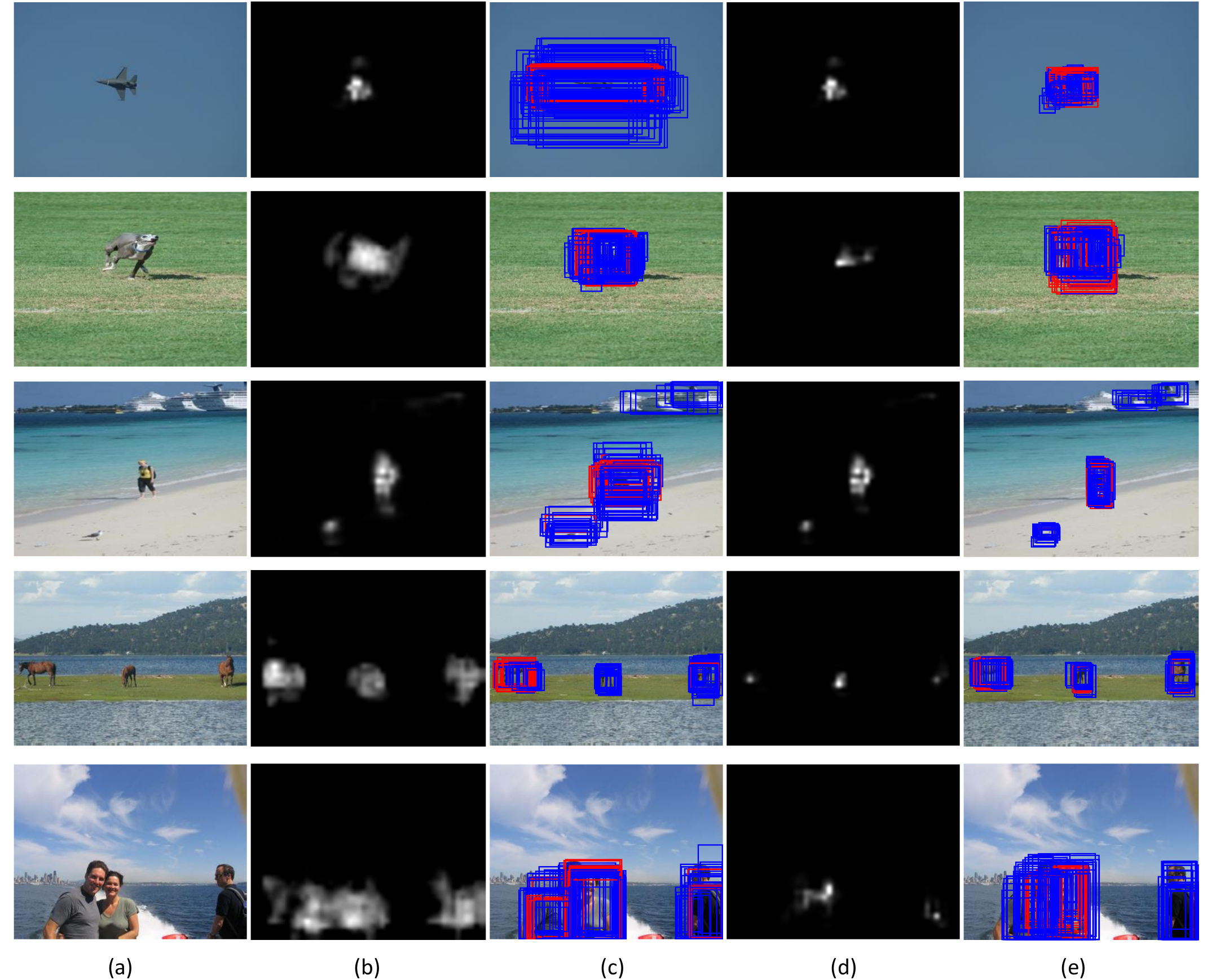}
		\end{center}
		\caption{Visualization of weights generated by ProbaNet in testing. The framework we utilize is Faster R-CNN. (a) Original images of the test dataset. (b) and (d): Visualization of weights in one feature map learned in ProbaNet corresponding to one kind of proposals. (c) and (e): Proposals mapping to original images with corresponding weights showed in (b) and (d). The blue proposals are assigned as top5$\%$ weights by ProbaNet and the red ones are assigned as top1$\%$ weights. (d) and (e) just as the same as (b) and (c) respectively.}
		\label{fig:w}
	\end{figure}

\begin{figure}
		\begin{center}
			\includegraphics[width=0.95\textwidth]{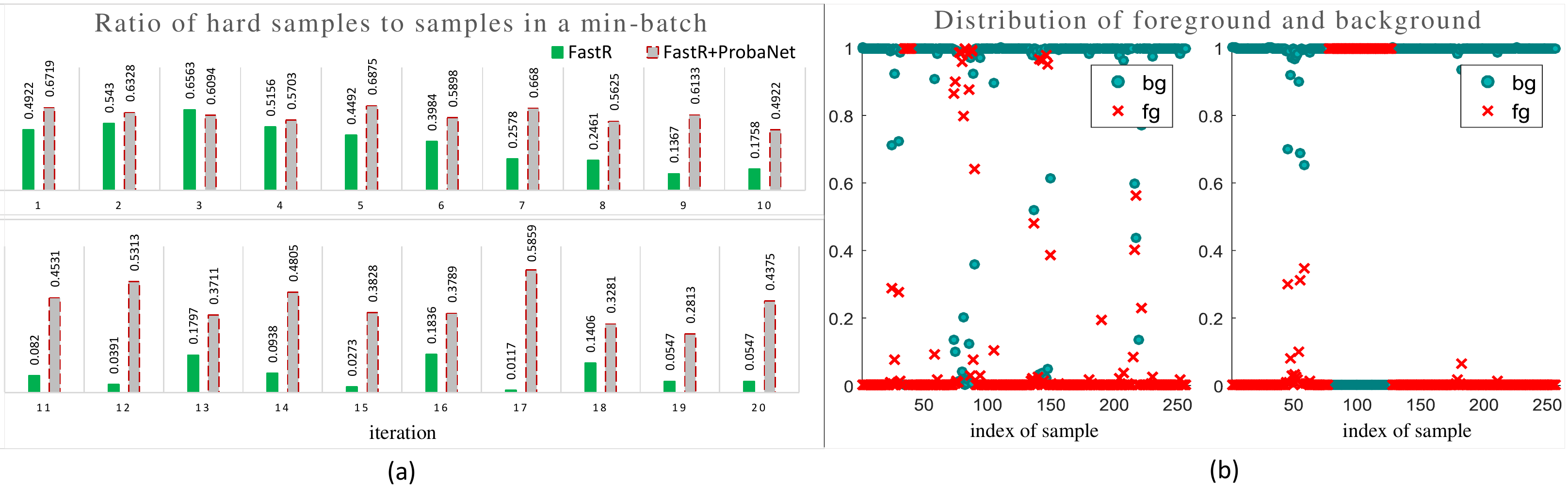}
		\end{center}
		\caption{(a) Comparison of hard proposals' ratio in a mini-batch between Faster R-CNN and our method in training, the horizontal axis refers the iteration in training. (b) Output after the softmax layer in RPN, the vertical axis and horizontal axis refer the output of softmax layer and the 256 samples in the mini-batch. The left one in (b) is results of Faster R-CNN, the right one is results of our method. The red fork in (b) is weight to be foreground and the blue circle is weight to be background.}
		\label{fig:rpn_out}
	\end{figure}

	Our approach is designed to assign appropriate weights to candidate proposals for helping counterweight easy-hard samples imbalance in object detection networks. Although we have shown that our method is effective from many aspects, to investigate the effectiveness of our method more deeply, we further do some experiments. To demonstrate the effectiveness of ProbaNet, in Figure \ref{fig:w} we visualize weights learned for candidate proposals and proposals with top 5$\%$ and top 1$\%$ weights generated by ProbaNet based RPN. In Figure \ref{fig:rpn_out} (a) we show the ratio of hard samples to all samples in a mini-batch during the training process. To validate the effectiveness of the statistical constraint in ProbaNet, we show the output after RPN with ProbaNet in Figure \ref{fig:rpn_out} (b).

	\noindent
	\textbf{Experiments for Visualizing the Focus of ProbaNet.} Figure \ref{fig:w} shows some samples for visualization purposes. We randomly select five test images which are displayed in column (a). We normalize weights corresponding to 2 kinds of proposals (total 9) generated by ProbaNet as shown in column (b) and (d). In column (b) and (d), the value of weights are larger when the color in positions approaching white, and minimum if that color is black. Every weight is related to a part of the area near the point where the weight exist. For display deeply, we further map the proposals to the original images and draw proposals with top 5$\%$ weights in blue and top 1$\%$ weights in red. These proposals in column (c) and (e) are corresponding to weights in column (b) and (d) respectively. It is obvious that the minority objects in the whole images have more focus than the majority background due to the higher weight. Thus from Figure \ref{fig:w}, It shows that ProbaNet gives the foreground proposals more focus but less for background proposals. Otherwise, better candidate proposals generally have higher weights. This fact is clearly observed by the airplane of picture 1, the person on the beach of picture 3 and horses near the lake of picture 4.
	
	\noindent
	\textbf{Experiments for Validating the effectiveness of ProbaNet.} As shown in Figure \ref{fig:rpn_out} (a), when combined with ProbaNet, the detector indeed select more hard proposals in a mini-batch. As is shown in Figure \ref{fig:rpn_out} (b), the results of our method have more discrimination capability. ProbaNet indeed alleviates the imbalance problem, the statistical constraint in ProbaNet indeed increase the degree of distinction among samples.
	
	\section{Conclusion}
	\label{sec:con}
	Object detectors based on CNNs face easy-hard samples imbalance problem, this issue has an adverse effect on performance. In this work, we proposed ProbaNet to combine with object detectors. ProbaNet not only alleviates the imbalance issue by suppressing the vast easy negatives in training stages but also emphasizes the minority but meaningful samples in testing. Experimental results confirmed that ProbaNet alleviates the imbalance problem, and improves the detection accuracy while offering less computational cost. 
	
	In the future, we will design adaptive adjustment parameters to replace the reduction parameter $r$ and the threshold $th$. Furthermore, we will apply the proposed ProbaNet to one-stage detectors for evaluating its performance and study imbalance problem more deeply and systematically.
	
	\bibliography{probanet}
    \bibliographystyle{probanet}

\end{document}